\documentclass{article}

\usepackage[preprint]{neurips_2025}
\usepackage[utf8]{inputenc}             
\usepackage[T1]{fontenc}                
\usepackage[hidelinks]{hyperref}
\usepackage{url}
\usepackage{booktabs}
\usepackage{nicefrac}
\usepackage{microtype}
\usepackage{amsfonts, amsmath}
\usepackage{amssymb}
\usepackage{graphicx}
\usepackage{xcolor}
\usepackage{wrapfig}
\usepackage{subcaption}
\usepackage[export]{adjustbox}
\usepackage{multirow}
\newcommand{\myspace}{\mathrel{\hspace{0.5em}}}
\definecolor{Y}{HTML}{FFE6CC}
\definecolor{P}{HTML}{E1D5E7}
\usepackage{threeparttable}
\usepackage{ragged2e}
\usepackage{etoolbox}
\usepackage{array}
\usepackage{comment}

\newcommand{\modelname}{FLAME}

\makeatletter
\@ifpackageloaded{natbib}{%
  \renewcommand\NAT@sort{1}%
}{%
  \PassOptionsToPackage{sort}{natbib}%
}%
\makeatother
\AtBeginEnvironment{tabular}{
  \setlength{\tabcolsep}{0pt}
}
\usepackage{enumitem}
\usepackage{xspace}
\usepackage{tabularx}
\usepackage{booktabs}
\usepackage{arydshln}

\newenvironment{packed_item}{
	\begin{itemize}[leftmargin=8pt,topsep=0pt,partopsep=0pt]
		\setlength{\itemsep}{0pt}
		\setlength{\parskip}{0pt}
		\setlength{\parsep}{0pt}
		\setlength{\topsep}{0pt}
	}
	{\end{itemize}}

\title{
    FLAME: Towards Federated Fine-Tuning Large Language Models Through Adaptive SMoE
}
\author{
Khiem Le$^1$, Tuan Tran$^2$, Ting Hua$^1$, Nitesh V. Chawla$^1$\\
$^1$ University of Notre Dame, IN, USA\\
$^2$ Trinity College Dublin, Ireland\\
\texttt{\{kle3, thua, nchawla\}@nd.edu}
}

\begin{document}
\maketitle

\begin{abstract}
Existing resource-adaptive LoRA federated fine-tuning methods enable clients to fine-tune models using compressed versions of global LoRA matrices, in order to accommodate various compute resources across clients. This compression requirement will lead to suboptimal performance due to information loss. To address this, we propose \modelname{}, a novel federated learning framework based on the Sparse Mixture-of-Experts (SMoE) architecture. Unlike prior approaches, \modelname{} retains full (uncompressed) global LoRA matrices and achieves client-side adaptability by varying the number of activated experts per client. However, incorporating SMoE into federated learning introduces unique challenges—specifically, the mismatch in output magnitude from partial expert activation and the imbalance in expert training quality across clients. \modelname{} tackles these challenges through a lightweight rescaling mechanism and an activation-aware aggregation scheme. Empirical results across diverse computational settings demonstrate that \modelname{} consistently outperforms existing methods, providing a robust and effective solution for resource-adaptive federated learning.
\end{abstract}

\section{Introduction}

Large language models (LLMs) have achieved remarkable success across a wide range of tasks, largely due to extensive pre-training on massive datasets  \cite{achiam2023gpt, dong2023towards, kelly2023bing, thirunavukarasu2023large}. While pre-trained LLMs demonstrate remarkable general abilities, fine-tuning remains essential for adapting these models to specific domains and applications \cite{anisuzzaman2025fine, parthasarathy2024ultimate}. However, centralizing sensitive data for fine-tuning raises significant privacy concerns \cite{voigt2017eu, pardau2018california} making federated learning (FL) an increasingly important paradigm that enables model improvement while preserving data locality \cite{zhang-etal-2023-fedpet, kuang2024federatedscope, woisetschlager2024federated, wu2025survey}.

Existing approaches \cite{cho-etal-2024-heterogeneous,bai2024federated}  typically maintain global LoRA matrices on central servers, while requiring each client to use a compressed version of the global LoRA matrices that aligns with its own computational constraints—typically by reducing the rank.  
These methods balance heterogeneous client capabilities primarily through matrix decomposition, particularly Singular Value Decomposition (SVD), operating under the assumption that higher-rank matrices preserve more information. 
Such a design deals with heterogeneity by allowing resource-constrained clients to use lower-rank approximations of the global LoRA matrices, while resource-rich clients can utilize higher-rank representations. 
Research has demonstrated that the importance ranking of singular values from SVD does not always align optimally with preserving LLM performance on downstream tasks \cite{hsulanguage,hua2022numerical}. 
Furthermore, this matrix decomposition-based strategy has inherent limitations: in order to accommodate diverse client capabilities, all local models are inherently forced to discard part of the knowledge encoded in the global LoRA matrices, which is obviously suboptimal.
Besides, our thorough investigation reveals crucial limitations of these methods. A deeper examination of FLOPs in our evaluation reveals that fine-tuning with LoRA matrices of smaller ranks does not remarkably reduce computational loads, since the computational demands for the base forward pass remain unchanged. This indicates that existing methods fundamentally fail to enable clients to complete fine-tuning using computational loads truly tailored to their resource budgets, representing a misleading direction in resource-adaptive federated learning.

\begin{figure}[!t]
    \centering
    \includegraphics[width=0.9\linewidth]{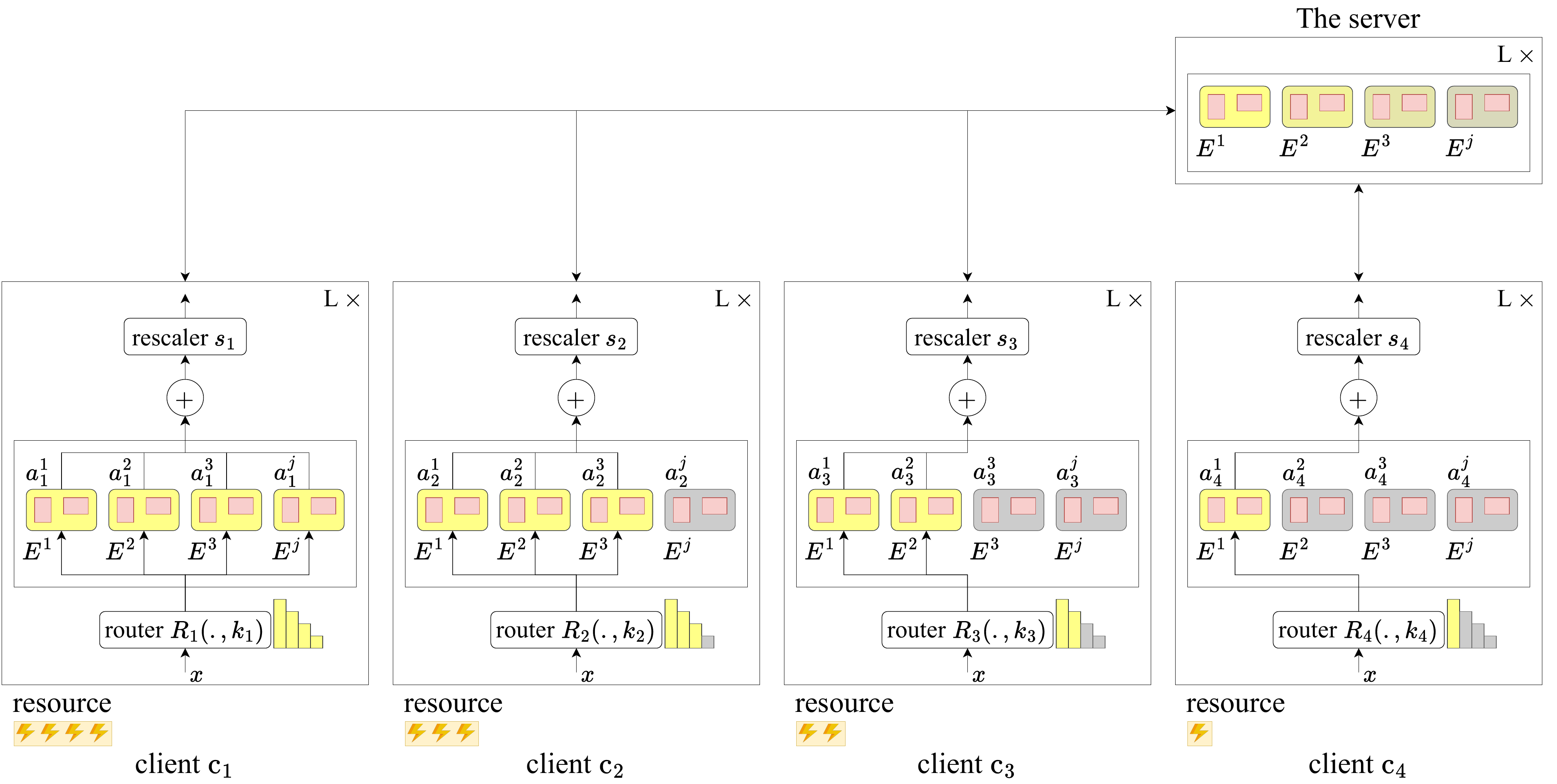}
    \caption{An illustration of resource-adaptive federated fine-tuning with \modelname.}
    \label{fig-FLAME}
\end{figure}

To address these limitations, we revisit the Sparse Mixture-of-Experts (SMoE) architecture and propose FLAME (\textbf{F}ederated \textbf{L}earning with \textbf{A}daptive sparse \textbf{M}ixture-of-\textbf{E}xperts), a novel federated learning framework that leverages SMoE to enable genuine resource-adaptive fine-tuning.
 As illustrated in Figure \ref{fig-FLAME}, FLAME allows each client to fine-tune using full (uncompressed) global LoRA matrices while varying the number of activated experts based on available compute capacity. Importantly, our examination of FLOPs verifies that this approach remarkably reduces computational loads since the base forward pass costs are accordingly reduced. In contrast to existing methods, FLAME truly enables clients to complete fine-tuning using computational loads tailored to their resource budgets. Furthermore, FLAME's approach offers significant advantages during deployment. By fine-tuning the model with global LoRA matrices while using smaller numbers of activated experts in SMoE layers, FLAME facilitates deploying the model with reduced expert activation during inference, thereby significantly enhancing deployment efficiency.

However, this design introduces two key challenges. First, partial expert activation creates output magnitude mismatches compared to full-capacity execution. \modelname{} addresses this through a lightweight learnable rescaling mechanism that adaptively calibrates outputs across different activation patterns. Second, an imbalance in expert activation frequency across clients results in an imbalance in the quality of their trained LoRA matrices, which can distort the global model after aggregation. To solve this, \modelname{} incorporates an activation-aware federated averaging scheme that incorporates activation frequency when generating global LoRA matrices, ensuring proper weighting of client contributions for each expert.
Our contributions can be summarized as follows:
\begin{packed_item}
    \item \textbf{Limitation analysis of existing methods:} We thoroughly investigate resource-adaptive federated fine-tuning LLMs and identify crucial limitations of existing methods, particularly their failure to enable true computational load adaptation tailored to client resource budgets.
    
    \item \textbf{A novel adaptive SMoE framework for federated learning:} We introduce \modelname, a novel federated learning framework that leverages sparse mixture-of-experts (SMoE) architecture to enable resource-adaptive fine-tuning without compromising the expressive power of global LoRA matrices. Unlike existing compression-based approaches, our method maintains full global LoRA matrices while varying the number of activated experts according to client computational capabilities.
    
    \item \textbf{Activation-aware aggregation scheme:} We develop an activation-aware federated averaging scheme that incorporates expert activation frequency across clients to generate balanced global LoRA matrices, addressing the shortcomings of standard federated averaging.
    
    \item \textbf{Comprehensive performance evaluation:} We demonstrate through extensive experiments on instruction-following tasks that \modelname{} achieves significantly better performance than existing methods across various computational settings and data distributions.
\end{packed_item}
\section{Methodology}


\subsection{Preliminaries}

We study fine-tuning large language models using Low-Rank Adaptation (LoRA) in a federated setting where data remains distributed across clients due to privacy concerns. A key challenge in this environment is accommodating the heterogeneous computational capabilities of participating devices through resource-adaptive federated fine-tuning.
Existing federated LoRA fine-tuning methods utilize a central server to coordinate training across clients. Each client holds a local dataset $D_i$ and operates under its resource constraint $\beta_i$. The server initializes and distributes global LoRA matrices of rank $r$, $\text{A} \in \mathbb{R}^{m \times r}$ and $\text{B} \in \mathbb{R}^{r \times n}$, for fine-tuning the frozen base model $\text{W} \in \mathbb{R}^{m \times n}$. Each client fine-tunes the LoRA matrices on its local data:
\begin{equation}
\text{A}_i = \text{A}, \quad \text{B}_i = \text{B},
\end{equation}
\begin{equation}
\min \frac{1}{|D_i|} \sum_{x \in D_i} \ell(\text{W}, x \mid \beta_i) \quad \text{s.t.} \quad h = \text{W}x + \text{A}_i \text{B}_i x.
\end{equation}

After local updates, clients return $\text{A}_i$ and $\text{B}_i$ to the server. Global LoRA matrices are then updated via federated averaging \cite{mcmahan2017communication, li2020on}. Each client's contribution is weighted by its dataset size $\gamma_i = |D_i|$, reflecting the \textit{training intensity} of its local updates:
\begin{equation}
\gamma_i = |D_i|, \quad \text{for } 1 \le i \le N,
\label{equ:gamma_original}
\end{equation}
\begin{equation}
\text{A} = \frac{\sum_{i=1}^{N} \gamma_i \text{A}_i}{\sum_{i=1}^{N} \gamma_i}, \quad 
\text{B} = \frac{\sum_{i=1}^{N} \gamma_i \text{B}_i}{\sum_{i=1}^{N} \gamma_i}.
\label{equ:AB_update_original}
\end{equation}

This framework ensures stable global aggregation by diminishing the impact of low-quality updates from clients with less data.


\begin{figure}[!t]
\centering
\begin{subfigure}[b]{0.495\textwidth}
\centering
\caption{AlpaGasus ($\alpha$ = 5)}
\includegraphics[width=0.900\textwidth]{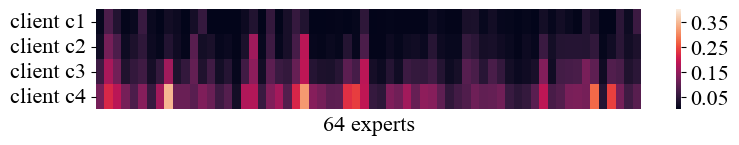}
\end{subfigure}%
\hfill
\begin{subfigure}[b]{0.495\textwidth}
\centering
\caption{AlpaGasus ($\alpha$ = 0.5)}
\includegraphics[width=0.900\textwidth]{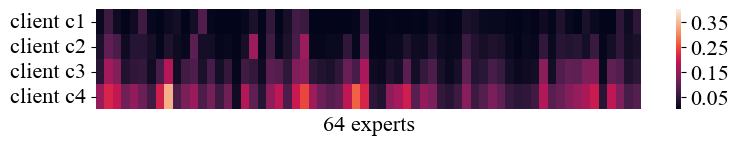}
\end{subfigure}%
\hfill
\begin{subfigure}[b]{0.495\textwidth}
\centering
\caption{Dolly ($\alpha$ = 5)}
\includegraphics[width=0.900\textwidth]{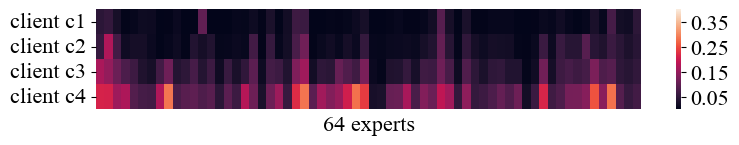}
\end{subfigure}%
\hfill
\begin{subfigure}[b]{0.495\textwidth}
\centering
\caption{Dolly ($\alpha$ = 0.5)}
\includegraphics[width=0.900\textwidth]{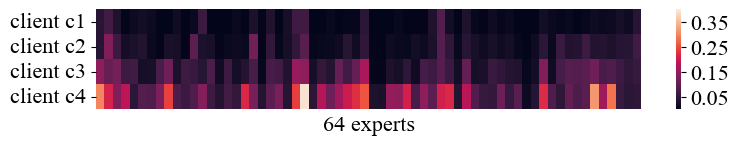}
\end{subfigure}%
\hfill
\caption{A highlight of the activation frequency of experts across clients in our experiments. The heatmaps display activation frequencies of all 64 experts (x-axis) across 4 clients (y-axis) for both AlpaGasus and Dolly datasets under different data heterogeneity settings.}
\label{fig-activation}
\end{figure}

\subsection{FLAME: Federated Learning with Adaptive MoE}

Our proposed FLAME considers an SMoE-based model. We denote $M$ matrices of parameters in SMoE layers as $\{\text{W}^j \in \mathbb{R}^{m \times n}\}_{j=1}^{M}$, corresponding to $M$ experts managed by a router that dynamically selects $k < M$ activated experts for processing input tokens based on routing scores. 

The federated learning procedure begins with a central server initializing and distributing experts' global LoRA matrices of rank $r$, $\{\text{A}^j \in \mathbb{R}^{m \times r}\}_{j=1}^{M}$ and $\{\text{B}^j \in \mathbb{R}^{r \times n}\}_{j=1}^{M}$, to all clients. Each client $c_i$, under resource constraint $\beta_i$, fine-tunes the model on its local dataset $D_i$ while adaptively using fewer activated experts ($k_i \leq k$) in SMoE layers:

\begin{equation*}
\{\text{A}_i^j\}_{j=1}^{M} = \{\text{A}^j\}_{j=1}^{M} \text{ and } \{\text{B}_i^j\}_{j=1}^{M} = \{\text{B}^j\}_{j=1}^{M},
\end{equation*}

\begin{equation}
\text{min} \frac{1}{|D_i|} \sum_{x \in D_i} \ell(\{\text{W}^j\}_{j=1}^{M}, x, k_i \text{ }|\text{ } \beta_i) \quad \text{s.t.} \quad h = s_i \cdot \sum_{j=1}^{M} R_i(x, k_i)^j \cdot (\text{W}^jx + \text{A}_i^j\text{B}_i^jx).
\end{equation}

Here, $R_i(x, k_i) = \text{TopK} (p(\{E^j\}_{j=1}^{M} | x), k_i)$ represents a router that selects $k_i$ activated experts for each input token by applying the $\text{TopK}$ function to routing probabilities. 
Since the router selects fewer experts than in the original configuration, the SMoE output diverges from the full-expert output, similar to the effect seen in dropout mechanisms \cite{srivastava2014dropout, labach2019survey}. 
To address this, FLAME incorporates a learnable rescaler $s_i \in \mathbb{R}$ that is trained to realign the output, rather than using a naive static ratio $\frac{k}{k_i}$. This adaptive approach better accounts for the dynamic nature of SMoE layers.

After local training, the server aggregates experts' trained LoRA matrices from clients to synthesize updated global LoRA matrices. However, standard federated averaging would be ineffective in this SMoE setting due to several factors:

\begin{packed_item}
\item The router $R_i(x, k_i)$ dynamically selects only $k_i \leq k$ activated experts for processing each input token, so only these selected experts are involved in any given training step;
\item Different experts receive different amounts of training across clients due to the routing mechanism;
\item Consequently, throughout $S_i$ training steps, a client's local dataset size alone no longer accurately reflects the \textit{training intensities} that each expert receives, and therefore no longer reliably indicates the quality of each expert's LoRA updates.
\end{packed_item}

To address these challenges, FLAME introduces an activation-aware aggregation scheme. This approach adjusts the weight of each expert's LoRA matrices by incorporating both the client's dataset size and the expert's activation frequency. This method better reflects the training intensity each expert has received, enabling the system to appropriately diminish the influence of low-quality expert updates while preserving high-quality ones, and therefore ensuring stable global LoRA matrices:

\begin{equation}
\{\gamma_i^j\}_{j=1}^{M} = \{(\frac{a_i^j}{S_i})^t \cdot |D_i|\}_{j=1}^{M} \text{ for }1 \leq i \leq N,
\label{equ:gamma_flame}
\end{equation}

\begin{equation}
\{\text{A}^j\}_{j=1}^{M} = \{\frac{\sum_{i=1}^{N} \gamma_i^j \text{A}_i^j}{\sum_{i=1}^{N} \gamma_i^j}\}_{j=1}^{M} \text{ and } \{\text{B}^j\}_{j=1}^{M} = \{\frac{\sum_{i=1}^{N} \gamma_i^j \text{B}_i^j}{\sum_{i=1}^{N} \gamma_i^j}\}_{j=1}^{M}.
\label{equ:AB_update_flame}
\end{equation}

Here, \( \frac{a_i^j}{S_i} \in [0, 1] \) denotes the activation frequency of expert \( j \) at client \( \text{c}_i \), computed as the number of times it was activated during \( S_i \) training steps. The temperature hyperparameter \( t \in \mathbb{N} \) adjusts the influence of activation frequency in computing aggregation weights. This design accounts for the high variance in expert usage across clients, as shown in Figure~\ref{fig-activation}.
This visualization empirically validates our hypothesis that expert activation is highly imbalanced in federated SMoE settings. 
The activation frequencies exhibit significant variation across experts and clients, with some experts being activated much more frequently (brighter colors) than others (darker regions). 
These observations directly motivate our activation-aware aggregation scheme in Equations \ref{equ:gamma_flame} and \ref{equ:AB_update_flame}, which weights each client's contribution to an expert's global parameters based on how frequently that client activated the expert during training. This approach ensures that clients with more experience training a particular expert have proportionally greater influence on that expert's final parameters, producing more stable and higher-quality global LoRA matrices.

\section{Evaluation}
\label{sec:evaluation}

\begin{table}[!t]
\centering
\caption{Evaluation setting up for resource-adaptive federated fine-tuning. $P$ and $\hat{P}$ denote total parameters and trainable parameters, respectively. $P_a$ and $\hat{P}_a$ denote active parameters and active trainable parameters, respectively. FLOPs are examined for a context of 128 input tokens.}
\label{tab-settings}
\scriptsize
\renewcommand{\arraystretch}{1.1}
\setlength\tabcolsep{0pt}
\begin{tabular*}{\linewidth}{@{\extracolsep{\fill}} lrrrrrrrrrrrrrrr }
\midrule
\multirow[t]{3}{*}{} &\multicolumn{4}{c}{OLMo-1.3B}                &\multicolumn{10}{c}{OLMoE-1.3B/6.9B}                                       \\
\cmidrule(l{0.00em}r{0.00em}){2-5}\cmidrule(l{0.00em}r{0.00em}){6-15}
                     &\multicolumn{4}{c}{HLoRA | FlexLoRA}  &\multicolumn{5}{c}{HLoRA | FlexLoRA} &\multicolumn{5}{c}{\modelname} \\
\cmidrule(l{0.00em}r{0.00em}){2-5}\cmidrule(l{0.00em}r{0.00em}){6-10}\cmidrule(l{0.00em}r{0.00em}){11-15}
          &$r$  &$P_a$/ $P$              &$\hat{P}_a$/ $\hat{P}$ &FLOPs                &$r$ &$k$ &$P_a$/ $P$              &$\hat{P}_a$/ $\hat{P}$ &FLOPs                &$r$ &$k$ &$P_a$/ $P$              &$\hat{P}_a$/ $\hat{P}$       &FLOPs \\
\cmidrule(l{0.00em}r{0.00em}){2-5}\cmidrule(l{0.00em}r{0.00em}){6-10}\cmidrule(l{0.00em}r{0.00em}){11-15}
$\beta_4$ &12   &\colorbox{Y}{1.3B}/1.3B &\colorbox{Y}{\phantom{1}9M}/\phantom{1}9M    &\colorbox{P}{337.2B} & 6  &8   &\colorbox{Y}{1.3B}/6.9B &\colorbox{Y}{\phantom{1}9M}/\phantom{1}58M   &\colorbox{P}{337.2B} &20  &1   &\colorbox{Y}{0.6B}/6.9B &\colorbox{Y}{\phantom{1}9M}/198M        &\colorbox{P}{158.0B (\phantom{1}46.1\%)} \\
$\beta_3$ &16   &\colorbox{Y}{1.3B}/1.3B &\colorbox{Y}{12M}/12M  &\colorbox{P}{338.0B} & 8  &8   &\colorbox{Y}{1.3B}/6.9B &\colorbox{Y}{12M}/\phantom{1}78M  &\colorbox{P}{338.0B} &20  &2   &\colorbox{Y}{0.7B}/6.9B &\colorbox{Y}{12M}/198M       &\colorbox{P}{184.4B (\phantom{1}53.8\%)} \\
$\beta_2$ &24   &\colorbox{Y}{1.3B}/1.3B &\colorbox{Y}{18M}/18M  &\colorbox{P}{339.6B} &12  &8   &\colorbox{Y}{1.3B}/6.9B &\colorbox{Y}{18M}/118M &\colorbox{P}{339.6B} &20  &4   &\colorbox{Y}{0.9B}/6.9B &\colorbox{Y}{18M}/198M       &\colorbox{P}{237.2B (\phantom{1}69.2\%)} \\
$\beta_1$ &40   &\colorbox{Y}{1.3B}/1.3B &\colorbox{Y}{30M}/30M  &\colorbox{P}{342.8B} &20  &8   &\colorbox{Y}{1.3B}/6.9B &\colorbox{Y}{30M}/198M &\colorbox{P}{342.8B} &20  &8   &\colorbox{Y}{1.3B}/6.9B &\colorbox{Y}{30M}/198M       &\colorbox{P}{342.8B (100.0\%)} \\
\midrule
\end{tabular*}
\end{table}


We evaluate resource-adaptive federated fine-tuning across two different model architectures: a dense model (OLMo-1.3B) and a sparse mixture-of-experts model (OLMoE-1.3B/6.9B). We define four resource configurations ($\beta_1$-$\beta_4$), representing decreasing parameter budgets.
We evaluate on two instruction-following datasets: AlpaGasus~\cite{chen2024alpagasus} (9K examples) and Dolly~\cite{conover2023free} (15K examples), using an 80\%/10\%/10\% split for training, validation, and testing.
Detailed experimental procedures, hyperparameters, and additional configurations are provided in Appendix~\ref{appendix:experiment-setup}.

\subsection{Matrix Compression Fails: A FLOPs-Based Comparison}
Table~\ref{tab-settings} presents our evaluation setup for resource-adaptive federated fine-tuning across different model architectures. We compare FLAME with existing matrix compression methods (HLoRA and FlexLoRA) using both dense (OLMo-1.3B) and sparse MoE (OLMoE-1.3B/6.9B) models. Four resource configurations ($\beta_1$-$\beta_4$) represent decreasing parameter budgets.

The table highlights a crucial finding: existing methods that reduce LoRA rank fail to meaningfully decrease computational requirements. For both model types, reducing LoRA ranks from configuration $\beta_1$ to $\beta_4$ only decreases FLOPs from 342.8B to 337.2B—a negligible 1.6\% reduction. This confirms our assertion that rank compression approaches fundamentally fail to enable true computational adaptation.
In contrast, FLAME maintains a constant LoRA rank ($r=20$) while reducing activated experts from 8 to 1 across configurations. This novel approach achieves the same parameter reduction targets while dramatically cutting computational costs—from 342.8B to 158.0B FLOPs (a 53.9\% reduction). This substantial difference in computational efficiency demonstrates why FLAME represents a fundamentally more effective approach to resource-adaptive federated fine-tuning.

\subsection{Performance Across Resource Budgets}

Table~\ref{tab:budgets-4-clients} presents comprehensive performance comparisons across different resource budgets. We conduct experiments with 4 clients, distributing the training data using Dirichlet distributions with concentration parameters $\alpha = \{5, 0.5\}$ to create varying degrees of data heterogeneity (higher $\alpha$ indicates more uniform distribution, lower $\alpha$ creates more skewed distributions). Each client is uniformly assigned one of the four resource budget configurations ($\beta_1$-$\beta_4$).
Our results show that FLAME consistently outperforms all baselines across all experimental settings:

\begin{packed_item}
    \item \textbf{Performance at constrained budgets:} At the most resource-constrained setting ($\beta_4$, 153.6B FLOPs), FLAME significantly outperforms all alternatives. For example, on AlpaGasus with $\alpha=5$, FLAME achieves a score of \textbf{24.14}, substantially exceeding the trivial OLMoE baseline (14.43), HLoRA (10.46), and FlexLoRA (12.20).
    
    \item \textbf{Performance across data distributions:} FLAME maintains its advantage across different data heterogeneity levels. For instance, on Dolly with $\alpha=0.5$ (higher heterogeneity), FLAME achieves \textbf{24.78} at $\beta_4$, while the best competing method achieves only 11.33.
    
    \item \textbf{Consistent superiority:} Even at higher resource budgets ($\beta_1$-$\beta_3$), FLAME consistently outperforms all other methods on both datasets and heterogeneity settings.
\end{packed_item}

Notably, existing methods (HLoRA and FlexLoRA) underperform even the trivial baseline (which simply employs a globally small LoRA rank for all experts) in the MoE setting. This underscores the ineffectiveness of simple rank compression strategies for SMoE models and highlights the advantage of \modelname's approach of adaptively reducing activated experts while maintaining LoRA rank.

\begin{table}[!t]
\centering
\caption{Performance comparison across different resource budgets.}
\label{tab:budgets-4-clients}
\scriptsize
\setlength\tabcolsep{0pt}
\begin{tabular*}{\linewidth}{@{\extracolsep{\fill}} lccccccccccccccc }
\midrule
\multirow[t]{3}{*}{\phantom{50\%}} &\multirow[t]{2}{*}{} &\multicolumn{7}{c}{AlpaGasus ($\alpha$ = 5)} &\multicolumn{7}{c}{AlpaGasus ($\alpha$ = 0.5)} \\
\cmidrule(l{0.00em}r{0.00em}){3-9}\cmidrule(l{0.00em}r{0.00em}){10-16}
& &\multicolumn{3}{c}{OLMo-1.3B} &\multicolumn{4}{c}{OLMoE-1.3B/6.9B} &\multicolumn{3}{c}{OLMo-1.3B} &\multicolumn{4}{c}{OLMoE-1.3B/6.9B} \\
\cmidrule(l{0.00em}r{0.00em}){3-5}\cmidrule(l{0.00em}r{0.00em}){6-9}\cmidrule(l{0.00em}r{0.00em}){10-12}\cmidrule(l{0.00em}r{0.00em}){13-16}
& FLOPs &trivial &HLoRA &FlexLoRA &trivial &HLoRA &FlexLoRA &\modelname &trivial &HLoRA &FlexLoRA &trivial &HLoRA &FlexLoRA &\modelname \\
\midrule
$\beta_4$ &153.6B &- &- &- &\underline{14.43} &10.46 &12.20 &\textbf{24.14} &- &- &- &\underline{13.60} &10.63 &10.66 &\textbf{24.88} \\
$\beta_3$ &179.2B &- &- &- &\underline{26.67} &24.52 &26.34 &\textbf{31.64} &- &- &- &\underline{27.45} &23.75 &25.64 &\textbf{31.84} \\
$\beta_2$ &230.4B &- &- &- &\underline{33.67} &32.32 &32.71 &\textbf{35.62} &- &- &- &\underline{33.32} &32.34 &33.28 &\textbf{35.03} \\
$\beta_1$ &332.8B &32.55 &31.59 &32.45 &\underline{36.60} &34.16 &34.41 &\textbf{36.63} &32.95 &31.24 &31.66 &\underline{35.54} &34.32 &34.93 &\textbf{37.39} \\
\midrule
\midrule
\multirow[t]{3}{*}{\phantom{50\%}} &\multirow[t]{2}{*}{} &\multicolumn{7}{c}{Dolly ($\alpha$ = 5)} &\multicolumn{7}{c}{Dolly ($\alpha$ = 0.5)} \\
\cmidrule(l{0.00em}r{0.00em}){3-9}\cmidrule(l{0.00em}r{0.00em}){10-16}
& &\multicolumn{3}{c}{OLMo-1.3B} &\multicolumn{4}{c}{OLMoE-1.3B/6.9B} &\multicolumn{3}{c}{OLMo-1.3B} &\multicolumn{4}{c}{OLMoE-1.3B/6.9B} \\
\cmidrule(l{0.00em}r{0.00em}){3-5}\cmidrule(l{0.00em}r{0.00em}){6-9}\cmidrule(l{0.00em}r{0.00em}){10-12}\cmidrule(l{0.00em}r{0.00em}){13-16}
& FLOPs &trivial &HLoRA &FlexLoRA &trivial &HLoRA &FlexLoRA &\modelname &trivial &HLoRA &FlexLoRA &trivial &HLoRA &FlexLoRA &\modelname \\
\midrule
$\beta_4$ &153.6B &- &- &- &\underline{10.79} &08.67 &09.70 &\textbf{26.74} &- &- &- &10.66 &09.17 &\underline{11.33} &\textbf{24.78} \\
$\beta_3$ &179.2B &- &- &- &\underline{25.78} &22.90 &24.02 &\textbf{32.22} &- &- &- &\underline{26.62} &24.80 &25.45 &\textbf{30.17} \\
$\beta_2$ &230.4B &- &- &- &\underline{34.99} &33.00 &33.75 &\textbf{36.89} &- &- &- &\underline{35.08} &32.91 &34.00 &\textbf{37.06} \\
$\beta_1$ &332.8B &32.76 &31.00 &31.74 &\underline{36.20} &35.22 &35.77 &\textbf{36.82} &32.40 &30.30 &31.08 &\underline{36.88} &34.87 &35.92 &\textbf{38.89} \\
\midrule
\end{tabular*}
\end{table}

\subsection{Performance with Larger Client Populations}

To validate FLAME's effectiveness in larger federated learning environments, we extend our experiments to cases with 40 clients, as presented in Table~\ref{tab:budgets-40-clients}. The datasets are distributed to 40 clients using Dirichlet distributions with concentration parameters $\alpha = \{5, 0.5\}$ to create varying degrees of data heterogeneity. The four resource budget configurations ($\beta_1$-$\beta_4$) are assigned uniformly across the client population. The results with 40 clients strongly reinforce our previous findings:

\begin{packed_item}
    \item \textbf{Consistent performance advantage:} FLAME maintains its superior performance across all settings, with particularly pronounced advantages at lower resource budgets. For example, on Dolly with $\alpha=0.5$ and the most constrained budget ($\beta_4$), FLAME achieves \textbf{21.16}, while the best competing method reaches only 8.53.
    
    \item \textbf{Scalability to larger client populations:} FLAME's performance advantage is maintained or even enhanced when scaling from 4 to 40 clients, demonstrating the robustness of our approach in larger federated learning scenarios.
    
    \item \textbf{Persistent pattern with rank-compression methods:} Similar to our observations with 4 clients, existing rank-compression methods (HLoRA, FlexLoRA) frequently underperform even the trivial baseline when applied to SMoE models. For instance, on AlpaGasus with $\alpha=5$ at budget $\beta_4$, the trivial baseline achieves 10.65, while HLoRA and FlexLoRA achieve only 9.54 and 9.29, respectively.
\end{packed_item}

These results from a scaled-up scenario with 40 clients further confirm the fundamental advantages of FLAME's expert-reduction approach over traditional rank-compression methods for resource-adaptive federated fine-tuning of SMoE models.

\subsection{Performance Under Client Sampling}

In practical federated learning environments, clients often have limited or intermittent availability. To evaluate performance under these realistic conditions, we conduct experiments with client sampling in our 40-client setup. Specifically, we randomly select only a subset of clients (participation rates of $p = \{50\%, 25\%\}$) to participate in each federated learning iteration.
Table~\ref{tab-40-50-25} presents extensive results under these client sampling scenarios. The findings demonstrate:

\begin{packed_item}
    \item \textbf{Consistent performance advantage:} FLAME maintains its superior performance across all sampling rates, datasets, and resource budgets. For example, with 50\% client participation on Dolly ($\alpha=0.5$) at budget $\beta_4$, FLAME achieves \textbf{17.52}, while the best alternative achieves only 7.15.
    
    \item \textbf{Enhanced robustness to reduced participation:} FLAME exhibits greater resilience to decreased client participation compared to all other methods. When participation drops from 100\% to 25\%, FLAME's performance degradation is less severe than that of competing approaches. For instance, on AlpaGasus ($\alpha=5$) at budget $\beta_4$, FLAME's performance decreases from 21.29 (100\% participation) to 19.29 (50\% participation) to 16.84 (25\% participation) - a more gradual decline than other methods experience.
    
    \item \textbf{Practical advantages in constrained settings:} At the most restrictive resource budget ($\beta_4$) and lowest client participation rate (25\%), FLAME still substantially outperforms all alternatives. For example, on Dolly ($\alpha=0.5$) with $\beta_4$ and 25\% participation, FLAME achieves \textbf{16.31}, more than double the performance of any competing method.
\end{packed_item}

These results highlight FLAME's exceptional resilience in practical federated learning scenarios with intermittent client availability. While all methods experience some performance degradation with reduced client participation, FLAME maintains a significant performance edge and degrades more gracefully than existing approaches. This robustness makes FLAME particularly well-suited for real-world federated learning deployments where client availability cannot be guaranteed.

\begin{table}[!t]
\centering
\caption{Performance comparison with 40 clients under different data distributions.}
\label{tab:budgets-40-clients}
\scriptsize
\setlength\tabcolsep{0pt}
\begin{tabular*}{\linewidth}{@{\extracolsep{\fill}} lccccccccccccccc }
\midrule
\multirow[t]{3}{*}{\phantom{50\%}} &\multirow[t]{2}{*}{} &\multicolumn{7}{c}{AlpaGasus ($\alpha$ = 5)} &\multicolumn{7}{c}{AlpaGasus ($\alpha$ = 0.5)} \\
\cmidrule(l{0.00em}r{0.00em}){3-9}\cmidrule(l{0.00em}r{0.00em}){10-16}
& &\multicolumn{3}{c}{OLMo-1.3B} &\multicolumn{4}{c}{OLMoE-1.3B/6.9B} &\multicolumn{3}{c}{OLMo-1.3B} &\multicolumn{4}{c}{OLMoE-1.3B/6.9B} \\
\cmidrule(l{0.00em}r{0.00em}){3-5}\cmidrule(l{0.00em}r{0.00em}){6-9}\cmidrule(l{0.00em}r{0.00em}){10-12}\cmidrule(l{0.00em}r{0.00em}){13-16}
& FLOPs &trivial &HLoRA &FlexLoRA &trivial &HLoRA &FlexLoRA &\modelname &trivial &HLoRA &FlexLoRA &trivial &HLoRA &FlexLoRA &\modelname \\
\midrule
$\beta_4$ &153.6B &- &- &- &\underline{10.65} &09.54 &09.29 &\textbf{21.29} &- &- &- &\underline{11.61} &09.97 &09.39 &\textbf{20.89} \\
$\beta_3$ &179.2B &- &- &- &\underline{24.08} &20.76 &22.20 &\textbf{29.11} &- &- &- &\underline{25.06} &20.23 &23.74 &\textbf{29.04} \\
$\beta_2$ &230.4B &- &- &- &\underline{32.33} &31.70 &31.76 &\textbf{33.74} &- &- &- &\underline{32.50} &31.65 &32.27 &\textbf{34.19} \\
$\beta_1$ &332.8B &30.46 &29.04 &29.92 &\underline{34.43} &34.22 &34.37 &\textbf{35.69} &31.29 &29.32 &29.98 &\underline{34.37} &33.84 &34.27 &\textbf{35.36} \\
\midrule
\midrule
\multirow[t]{3}{*}{\phantom{50\%}} &\multirow[t]{2}{*}{} &\multicolumn{7}{c}{Dolly ($\alpha$ = 5)} &\multicolumn{7}{c}{Dolly ($\alpha$ = 0.5)} \\
\cmidrule(l{0.00em}r{0.00em}){3-9}\cmidrule(l{0.00em}r{0.00em}){10-16}
& &\multicolumn{3}{c}{OLMo-1.3B} &\multicolumn{4}{c}{OLMoE-1.3B/6.9B} &\multicolumn{3}{c}{OLMo-1.3B} &\multicolumn{4}{c}{OLMoE-1.3B/6.9B} \\
\cmidrule(l{0.00em}r{0.00em}){3-5}\cmidrule(l{0.00em}r{0.00em}){6-9}\cmidrule(l{0.00em}r{0.00em}){10-12}\cmidrule(l{0.00em}r{0.00em}){13-16}
& FLOPs &trivial &HLoRA &FlexLoRA &trivial &HLoRA &FlexLoRA &\modelname &trivial &HLoRA &FlexLoRA &trivial &HLoRA &FlexLoRA &\modelname \\
\midrule
$\beta_4$ &153.6B &- &- &- &\underline{07.88} &07.73 &07.81 &\textbf{19.56} &- &- &- &\underline{08.53} &08.05 &07.01 &\textbf{21.16} \\
$\beta_3$ &179.2B &- &- &- &\underline{25.44} &19.06 &20.87 &\textbf{29.18} &- &- &- &\underline{24.38} &18.59 &22.15 &\textbf{30.54} \\
$\beta_2$ &230.4B &- &- &- &\underline{32.23} &30.13 &31.16 &\textbf{33.49} &- &- &- &\underline{32.96} &29.02 &30.98 &\textbf{34.30} \\
$\beta_1$ &332.8B &31.26 &29.37 &30.34 &\underline{35.11} &33.99 &34.20 &\textbf{35.25} &31.88 &28.90 &30.24 &\underline{34.94} &32.91 &34.32 &\textbf{34.98} \\
\midrule
\end{tabular*}
\end{table}

\begin{table}[!t]
\centering
\caption{Performance under client sampling with 40 clients.}
\label{tab-40-50-25}
\scriptsize
\setlength\tabcolsep{0pt}
\begin{tabular*}{\linewidth}{@{\extracolsep{\fill}} lccccccccccccccc }
\midrule
\multirow[t]{3}{*}{50\%} &\multirow[t]{2}{*}{} &\multicolumn{7}{c}{AlpaGasus ($\alpha$ = 5)} &\multicolumn{7}{c}{AlpaGasus ($\alpha$ = 0.5)} \\
\cmidrule(l{0.00em}r{0.00em}){3-9}\cmidrule(l{0.00em}r{0.00em}){10-16}
& &\multicolumn{3}{c}{OLMo-1.3B} &\multicolumn{4}{c}{OLMoE-1.3B/6.9B} &\multicolumn{3}{c}{OLMo-1.3B} &\multicolumn{4}{c}{OLMoE-1.3B/6.9B} \\
\cmidrule(l{0.00em}r{0.00em}){3-5}\cmidrule(l{0.00em}r{0.00em}){6-9}\cmidrule(l{0.00em}r{0.00em}){10-12}\cmidrule(l{0.00em}r{0.00em}){13-16}
& FLOPs &trivial &HLoRA &FlexLoRA &trivial &HLoRA &FlexLoRA &\modelname &trivial &HLoRA &FlexLoRA &trivial &HLoRA &FlexLoRA &\modelname \\
\midrule
$\beta_4$ &153.6B &- &- &- &\underline{11.11} &08.55 &08.28 &\textbf{19.29} &- &- &- &\underline{10.73} &08.35 &08.45 &\textbf{19.42} \\
$\beta_3$ &179.2B &- &- &- &\underline{23.02} &18.52 &20.47 &\textbf{26.53} &- &- &- &\underline{22.30} &18.06 &19.13 &\textbf{27.60} \\
$\beta_2$ &230.4B &- &- &- &\underline{29.49} &27.81 &29.12 &\textbf{32.28} &- &- &- &28.97 &28.27 &\underline{28.99} &\textbf{32.40} \\
$\beta_1$ &332.8B &27.79 &26.50 &26.93 &30.59 &\underline{30.70} &30.68 &\textbf{33.88} &27.49 &25.87 &26.97 &30.69 &30.26 &\underline{30.88} &\textbf{33.81} \\
\midrule
\midrule
\multirow[t]{3}{*}{50\%} &\multirow[t]{2}{*}{} &\multicolumn{7}{c}{Dolly ($\alpha$ = 5)} &\multicolumn{7}{c}{Dolly ($\alpha$ = 0.5)} \\
\cmidrule(l{0.00em}r{0.00em}){3-9}\cmidrule(l{0.00em}r{0.00em}){10-16}
& &\multicolumn{3}{c}{OLMo-1.3B} &\multicolumn{4}{c}{OLMoE-1.3B/6.9B} &\multicolumn{3}{c}{OLMo-1.3B} &\multicolumn{4}{c}{OLMoE-1.3B/6.9B} \\
\cmidrule(l{0.00em}r{0.00em}){3-5}\cmidrule(l{0.00em}r{0.00em}){6-9}\cmidrule(l{0.00em}r{0.00em}){10-12}\cmidrule(l{0.00em}r{0.00em}){13-16}
& FLOPs &trivial &HLoRA &FlexLoRA &trivial &HLoRA &FlexLoRA &\modelname &trivial &HLoRA &FlexLoRA &trivial &HLoRA &FlexLoRA &\modelname \\
\midrule
$\beta_4$ &153.6B &- &- &- &06.41 &\underline{07.16} &06.86 &\textbf{14.93} &- &- &- &06.25 &06.79 &\underline{07.15} &\textbf{17.52} \\
$\beta_3$ &179.2B &- &- &- &\underline{21.83} &17.60 &19.98 &\textbf{25.79} &- &- &- &\underline{22.64} &15.49 &17.51 &\textbf{26.66} \\
$\beta_2$ &230.4B &- &- &- &\underline{29.37} &26.30 &28.75 &\textbf{31.97} &- &- &- &\underline{29.68} &25.64 &28.29 &\textbf{32.58} \\
$\beta_1$ &332.8B &27.83 &26.69 &27.45 &\underline{30.80} &29.99 &30.65 &\textbf{34.59} &28.75 &26.53 &27.26 &\underline{31.86} &29.78 &30.90 &\textbf{34.73} \\
\midrule

\midrule
\multirow[t]{3}{*}{25\%} &\multirow[t]{2}{*}{} &\multicolumn{7}{c}{AlpaGasus ($\alpha$ = 5)} &\multicolumn{7}{c}{AlpaGasus ($\alpha$ = 0.5)} \\
\cmidrule(l{0.00em}r{0.00em}){3-9}\cmidrule(l{0.00em}r{0.00em}){10-16}
& &\multicolumn{3}{c}{OLMo-1.3B} &\multicolumn{4}{c}{OLMoE-1.3B/6.9B} &\multicolumn{3}{c}{OLMo-1.3B} &\multicolumn{4}{c}{OLMoE-1.3B/6.9B} \\
\cmidrule(l{0.00em}r{0.00em}){3-5}\cmidrule(l{0.00em}r{0.00em}){6-9}\cmidrule(l{0.00em}r{0.00em}){10-12}\cmidrule(l{0.00em}r{0.00em}){13-16}
& FLOPs &trivial &HLoRA &FlexLoRA &trivial &HLoRA &FlexLoRA &\modelname &trivial &HLoRA &FlexLoRA &trivial &HLoRA &FlexLoRA &\modelname \\
\midrule
$\beta_4$ &153.6B &- &- &- &\underline{11.12} &08.53 &08.86 &\textbf{16.84} &- &- &- &\underline{10.58} &08.31 &08.07 &\textbf{18.26} \\
$\beta_3$ &179.2B &- &- &- &\underline{22.83} &20.63 &21.34 &\textbf{26.28} &- &- &- &\underline{22.78} &18.03 &20.14 &\textbf{25.38} \\
$\beta_2$ &230.4B &- &- &- &\underline{28.82} &28.45 &28.68 &\textbf{30.73} &- &- &- &\underline{29.66} &28.07 &28.43 &\textbf{30.57} \\
$\beta_1$ &332.8B &26.79 &26.22 &26.86 &\underline{30.89} &30.25 &30.40 &\textbf{33.08} &27.44 &26.02 &26.72 &\underline{30.78} &30.13 &30.38 &\textbf{32.59} \\
\midrule
\midrule
\multirow[t]{3}{*}{25\%} &\multirow[t]{2}{*}{} &\multicolumn{7}{c}{Dolly ($\alpha$ = 5)} &\multicolumn{7}{c}{Dolly ($\alpha$ = 0.5)} \\
\cmidrule(l{0.00em}r{0.00em}){3-9}\cmidrule(l{0.00em}r{0.00em}){10-16}
& &\multicolumn{3}{c}{OLMo-1.3B} &\multicolumn{4}{c}{OLMoE-1.3B/6.9B} &\multicolumn{3}{c}{OLMo-1.3B} &\multicolumn{4}{c}{OLMoE-1.3B/6.9B} \\
\cmidrule(l{0.00em}r{0.00em}){3-5}\cmidrule(l{0.00em}r{0.00em}){6-9}\cmidrule(l{0.00em}r{0.00em}){10-12}\cmidrule(l{0.00em}r{0.00em}){13-16}
& FLOPs &trivial &HLoRA &FlexLoRA &trivial &HLoRA &FlexLoRA &\modelname &trivial &HLoRA &FlexLoRA &trivial &HLoRA &FlexLoRA &\modelname \\
\midrule
$\beta_4$ &153.6B &- &- &- &05.28 &06.97 &\underline{08.11} &\textbf{13.89} &- &- &- &06.76 &06.74 &\underline{07.19} &\textbf{16.31} \\
$\beta_3$ &179.2B &- &- &- &\underline{22.55} &18.24 &19.45 &\textbf{24.47} &- &- &- &\underline{21.37} &17.06 &21.06 &\textbf{24.85} \\
$\beta_2$ &230.4B &- &- &- &\underline{28.54} &26.70 &28.07 &\textbf{30.49} &- &- &- &28.13 &26.69 &\underline{28.37} &\textbf{31.25} \\
$\beta_1$ &332.8B &28.18 &26.53 &27.97 &30.38 &30.46 &\underline{30.56} &\textbf{33.33} &28.18 &26.71 &27.23 &\underline{31.58} &29.71 &30.00 &\textbf{33.14} \\
\midrule
\end{tabular*}
\end{table}

\begin{table}[!t]
\centering
\caption{Impact of the rescaler, results are reported on OLMoE-1.3B/6.9B.}
\label{tab-rescaler}
\scriptsize
\setlength\tabcolsep{0pt}
\begin{tabular*}{\linewidth}{@{\extracolsep{\fill}} lcccccccc}
\midrule
\multirow{3}{*}{} &\multirow{2}{*}{} &\multicolumn{3}{c}{AlpaGasus ($\alpha$ = 5)} &\multicolumn{3}{c}{AlpaGasus ($\alpha$ = 0.5)} \\
\cmidrule(l{0.00em}r{0.00em}){3-5}\cmidrule(l{0.00em}r{0.00em}){6-8}
\cmidrule(l{0.00em}r{0.00em}){3-5}\cmidrule(l{0.00em}r{0.00em}){6-8}
&FLOPs &No rescaler &static $k / k_i$ &$s_i$ (\modelname) &No rescaler &static $k / k_i$ &$s_i$ (\modelname) \\
\midrule
$\beta_4$ &153.6B &24.12 &23.78 &\textbf{24.14} &24.28 &24.04 &\textbf{24.88} \\
$\beta_3$ &179.2B &30.47 &30.13 &\textbf{31.64} &\textbf{31.88} &31.47 &31.84 \\
$\beta_2$ &230.4B &\textbf{35.96} &35.49 &35.62 &\textbf{36.60} &36.01 &35.03 \\
$\beta_1$ &332.8B &36.06 &35.48 &\textbf{36.63} &37.38 &36.74 &\textbf{37.39} \\
\midrule
\midrule
\multirow{3}{*}{} &\multirow{2}{*}{} &\multicolumn{3}{c}{Dolly ($\alpha$ = 5)} &\multicolumn{3}{c}{Dolly ($\alpha$ = 0.5)} \\
\cmidrule(l{0.00em}r{0.00em}){3-5}\cmidrule(l{0.00em}r{0.00em}){6-8}
\cmidrule(l{0.00em}r{0.00em}){3-5}\cmidrule(l{0.00em}r{0.00em}){6-8}
&FLOPs &No rescaler &static $k / k_i$ &$s_i$ (\modelname) &No rescaler &static $k / k_i$ &$s_i$ (\modelname) \\
\midrule
$\beta_4$ &153.6B &26.61 &26.32 &\textbf{26.74} &24.35 &24.03 &\textbf{24.78} \\
$\beta_3$ &179.2B &32.01 &31.43 &\textbf{32.22} &\textbf{31.37} &30.87 &30.17 \\
$\beta_2$ &230.4B &35.07 &34.44 &\textbf{36.89} &36.40 &35.82 &\textbf{37.06} \\
$\beta_1$ &332.8B &36.04 &35.43 &\textbf{36.82} &37.58 &36.94 &\textbf{38.89} \\
\midrule
\end{tabular*}
\end{table}

\subsection{Ablation Studies}


\textbf{Impact of the Rescaler}.
A critical component of \modelname{} is the learnable rescaler $s_i$ that helps normalize the outputs when varying numbers of experts are activated. To evaluate its importance, we compare three variants:
1) \textbf{\modelname{} with learnable rescaler} $s_i$: Our full approach with learned rescaling factors.
2) \textbf{Static rescaler} $k/k_i$: A deterministic rescaling based on the ratio between the standard number of activated experts $k$ and the client-specific reduced number $k_i$.
3) \textbf{No rescaler}: A variant without rescaling.
Table~\ref{tab-rescaler} presents these results across different datasets, data distributions, and resource budgets. Several important patterns emerge:
\begin{packed_item}
\item \textbf{Learnable rescaler advantage}: The learnable rescaler $s_i$ generally yields the best or highly competitive performance, outperforming the "No rescaler" variant in 10 out of 16 experimental settings. For example, on Dolly with $\alpha=0.5$ and budget $\beta_1$, \modelname{} with $s_i$ achieves \textbf{38.89}, compared to 37.58 without a rescaler.

\item \textbf{Static rescaler limitations}: The static rescaler ($k/k_i$) consistently underperforms other variants across nearly all settings. For instance, on AlpaGasus with $\alpha=5$ and budget $\beta_4$, the static rescaler achieves only 23.78, whereas the learnable rescaler achieves 24.14, and no rescaler achieves 24.12.

\item \textbf{Resource-dependent impact}: The advantage of the learnable rescaler becomes more pronounced at certain resource levels. At $\beta_3$ (179.2B FLOPs), the learnable rescaler consistently outperforms both alternatives across all datasets and distribution settings.
\end{packed_item}

These results demonstrate the learnable rescaler provides \modelname{} with more flexibility towards better performance, particularly in resource-constrained settings.

\begin{figure}[!t]
\centering
\begin{subfigure}[b]{0.48\textwidth}
\centering
\caption{AlpaGasus ($\alpha$ = 5)}
\includegraphics[width=0.975\textwidth]{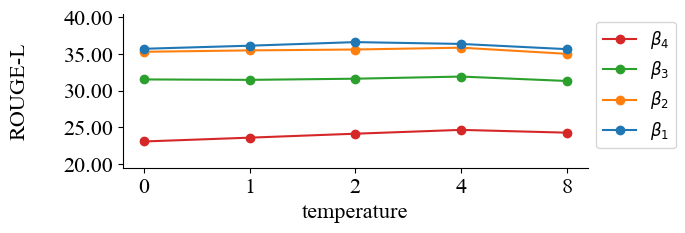}
\end{subfigure}%
\hfill
\begin{subfigure}[b]{0.48\textwidth}
\centering
\caption{AlpaGasus ($\alpha$ = 0.5)}
\includegraphics[width=0.975\textwidth]{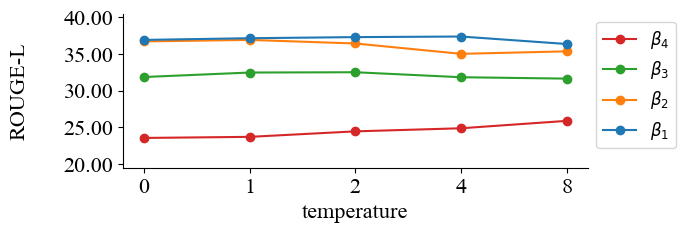}
\end{subfigure}%
\hfill
\begin{subfigure}[b]{0.48\textwidth}
\centering
\caption{Dolly ($\alpha$ = 5)}
\includegraphics[width=0.975\textwidth]{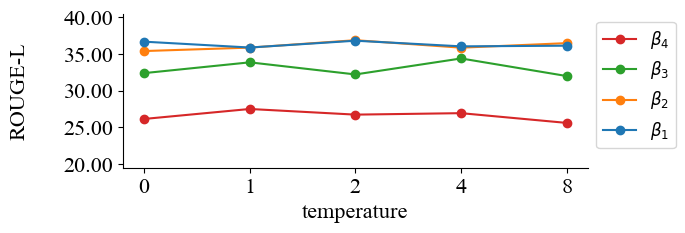}
\end{subfigure}%
\hfill
\begin{subfigure}[b]{0.48\textwidth}
\centering
\caption{Dolly ($\alpha$ = 0.5)}
\includegraphics[width=0.975\textwidth]{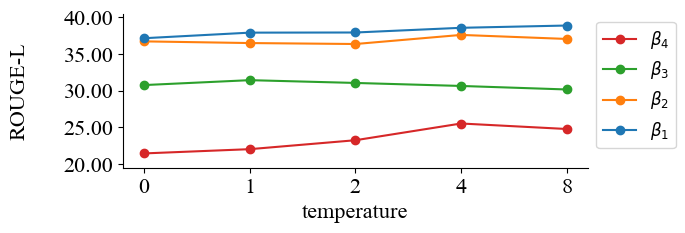}
\end{subfigure}%
\hfill
\caption{Impact of the temperature, results are reported on OLMoE-1.3B/6.9B. }
\label{fig-temperature}
\end{figure}

\textbf{Impact of Temperature in Activation-Aware Aggregation}. We examine how the temperature parameter $t$ in our activation-aware aggregation scheme affects model performance. This parameter controls how strongly the aggregation favors clients where an expert is frequently activated.
Figure~\ref{fig-temperature} presents performance results across different temperature values ranging from $t=0$ (equivalent to standard federated averaging) to $t=8$ (strongly favoring high-activation clients) for all datasets, data distributions, and resource budgets. Several key observations can be made:
\begin{packed_item}

\item \textbf{Resource-dependent temperature sensitivity}: The most resource-constrained setting ($\beta_4$, red lines) shows the highest sensitivity to temperature, with performance consistently improving as temperature increases up to $t=4$ or $t=8$. For example, on AlpaGasus ($\alpha=0.5$), performance at $\beta_4$ increases steadily from $t=0$ to $t=8$.

\item \textbf{Dataset-specific patterns}: The Dolly dataset with $\alpha=0.5$ (high heterogeneity) shows particularly strong improvements with higher temperatures at $\beta_4$, suggesting that activation-aware aggregation is especially beneficial for resource-constrained settings with heterogeneous data distributions.

\item \textbf{Benefit of activation-aware aggregation}: Across nearly all configurations, incorporating activation frequency ($t>0$) outperforms standard federated averaging ($t=0$). This confirms our hypothesis that experts should be more heavily influenced by clients where they are frequently activated.
    
\end{packed_item}

These findings validate our activation-aware aggregation approach and show that assigning higher weights to clients where experts are frequently activated leads to better performance. The results suggest that a temperature value of $t=2$ to $t=4$ strikes a good balance across most configurations, though the optimal value may depend on the specific resource constraints and data distributions.


\section{Related Work}

\textbf{Parameter-efficient Federated Fine-tuning for LLMs} represents a broad research area addressing federated learning with reduced parameter requirements. This area includes: (1) \textit{LoRA-based} methods that decompose weight updates into low-rank approximations, such as FeDeRA \cite{yan2024federa}, which uses SVD for initialization; FedSA-LoRA \cite{guo2024selective}, which selectively shares matrices; FFA-LoRA \cite{sun2024improving}, which freezes A matrices; and FEDHM \cite{yao2025fedhm}, which aggregates low-rank models into full-rank ones. (2) \textit{Prompt-based} approaches like PROMPTFL \cite{guo2023promptfl} and FedBPT \cite{sun2023fedbpt} optimize input prompts rather than model parameters. (3) \textit{Adapter-based} methods insert specialized modules between frozen layers, including FedAdapter \cite{cai2022fedadapter}, which adjusts adapter dimensions; FedTTT \cite{ghiasvand2024communication}, which uses tensor decomposition; and C2A \cite{kim2023client}, which generates client-specific adapters via hypernetworks.

\textbf{Resource-adaptive Federated Fine-tuning}, most relevant to our work, specifically addresses heterogeneous client computational capabilities within federated learning. Existing approaches include FedIT \cite{zhang2024towards}, which applied FedAvg \cite{mcmahan2017communication} to LoRA with fixed ranks; FLoRA \cite{wang2024flora}, introducing stacking-based aggregation; HLoRA \cite{cho-etal-2024-heterogeneous}, which distributes truncated LoRA modules and uses sparsity-weighted aggregation; and FlexLoRA \cite{bai2024federated}, which leverages SVD for dynamic rank adjustment based on client resources. These methods primarily focus on compressing global LoRA matrices at different levels to accommodate diverse client capabilities, which our work identifies as fundamentally limiting for achieving true computational load adaptation. 
\section{Discussion}
\label{sec:discussion}

Recall our activation-aware aggregation scheme shown in Equation 
\ref{equ:gamma_flame}.
To justify our design choices, we analyze the correlation between our activation-aware aggregation scheme and standard federated averaging through key edge cases:
\begin{packed_item}
\item \textbf{Temperature effect:} When the temperature $t$ is set to 0, the term $(\frac{a_i^j}{S_i})^t$ becomes 1 regardless of activation frequency, reducing our scheme to standard federated averaging.
\item \textbf{Full activation:} When expert $j$ is activated during all $S_i$ training steps at client $\text{c}_i$, $\frac{a_i^j}{S_i}$ equals 1. This gives full weight to that client's updates for this expert, equal to standard federated averaging.
\item \textbf{Zero activation:} If expert $j$ is never activated during training at client $\text{c}_i$, $\frac{a_i^j}{S_i}$ equals 0, resulting in zero contribution from that client for this expert. This scheme correctly prevents randomly initialized local LoRA matrices from contaminating the global model.
\end{packed_item}

\textbf{Limitations.} Our method is specifically designed for federated fine-tuning of SMoE-based LLMs. While this might appear restrictive, it aligns with industry trends, as most modern LLMs increasingly adopt SMoE architecture for scalability and efficiency \cite{liu2024deepseek, guo2025deepseek, meta2025llama, cai2025survey}. Due to computational resource constraints, our experiments were limited to OLMoE-1.3B/6.9B. 

\section{Conclusion}
We present FLAME, a novel federated fine-tuning framework based on sparse mixture-of-experts that enables true resource adaptivity without compressing global LoRA matrices.
Through our learnable rescaling scheme and activation-aware aggregation mechanism, FLAME consistently outperforms existing approaches across diverse setting. 
As LLM increasingly adopt SMoE architectures, FLAME offers a practical federated learning solution for democratizing access to powerful large language models while maintaining privacy protections for sensitive or resource-constrained environments.


\bibliographystyle{plain}
\bibliography{refs}

\clearpage
\appendix\renewcommand{\thesection}{A\arabic{section}}


\clearpage
\section{Experimental Setup}
\label{appendix:experiment-setup}
\subsection{Dense Model Experiments}
For the dense model, we use OLMo-1.3B~\cite{groeneveld2024olmo} with $P=P_a=1.3$B parameters. We implement existing methods (HLoRA and FlexLoRA) with LoRA ranks $r \in \{40, 24, 16, 12\}$ for configurations $\beta_1$-$\beta_4$, yielding trainable parameters $\hat{P}=\hat{P}_a \in \{30, 18, 12, 9\}$M. The computational cost remains relatively stable across configurations, from 342.8B FLOPs at $\beta_1$ to 337.2B FLOPs at $\beta_4$.

\subsection{Sparse MoE Experiments}
For the sparse MoE model, we use OLMoE-1.3B/6.9B~\cite{muennighoff2024olmoe}, which contains 64 experts per layer with $P=6.9$B total parameters, but only activates a subset of experts per token. For existing methods (HLoRA and FlexLoRA) on OLMoE, we maintain $k=8$ activated experts ($P_a=1.3$B) and create configurations with LoRA ranks $r \in \{20, 12, 8, 6\}$, yielding $\hat{P}_a \in \{30, 18, 12, 9\}$M active trainable parameters out of $\hat{P} \in \{198, 118, 78, 58\}$M total trainable parameters. Similar to the dense model, FLOPs remain stable from 342.8B to 337.2B across configurations.

For FLAME on OLMoE, we take a fundamentally different approach: we maintain a constant LoRA rank $r=20$ while reducing activated experts to $k \in \{8, 4, 2, 1\}$ across configurations $\beta_1$-$\beta_4$. This strategy maintains the same $\hat{P}_a \in \{30, 18, 12, 9\}$M active trainable parameters while progressively reducing active parameters $P_a \in \{1.3, 0.9, 0.7, 0.6\}$B. Crucially, this approach significantly reduces computational loads, from 342.8B FLOPs (100\%) in $\beta_1$ to only 158.0B FLOPs (46.1\%) in $\beta_4$.

\section{Implementation Details}
\label{appendix:implementation}
\subsection{FLOPs Profiling}

The Floating Point Operations (FLOPs) for our experiments were calculated using DeepSpeed's profiling tool \cite{10.1145/3394486.3406703}. To determine the computational cost attributable specifically to Low-Rank Adaptation (LoRA), we first measure the FLOPs of the base model. Subsequently, LoRA was applied to this base model, and the total FLOPs of the resulting unmerged LoRA-adapted model were measured. The FLOPs contributed by the LoRA components were then calculated as the difference between the total FLOPs of the LoRA-adapted model and the FLOPs of the original base model. 

For Mixture-of-Experts (MoE) models, this process was extended as follows: the base MoE model was initially configured with a specific number of top-$k$ active experts, and its FLOPs were profiled. LoRA was then applied to this top-$k$-configured MoE model, and its FLOPs were measured. The incremental FLOPs due to LoRA in the MoE context were similarly derived by subtracting the FLOPs of the top-$k$-configured base MoE model from those of the LoRA-enhanced MoE model. All FLOPs measurements were conducted using sample input tensors generated with sequence lengths of $128$ and batch sizes of $1$. 

\subsection{Hardware and Hyper-parameters}
All of our experiments were conducted using two NVIDIA A100 80GB GPUs. For approaches utilizing LoRA, we use the scaling parameter \(\alpha\) of 16. On the client side, fine-tuning is conducted with the Adam optimizer with a learning rate of \(1.5 \times 10^{-4}\) and a batch size of 16. Each client conducts local training for a single epoch in every communication round. The overall federated training process consists of 2 communication rounds between clients and the central server.

\subsection{Prompt Template}

In our experiments, data instances are wrapped to prompts before processing by LLMs. We directly apply the template provided by Alpaca to the datasets in our experiments. For better reproducibility, we present how we fill the fields in the template with the attributes of data instances in Table \ref{tab:prompt_template}. 

\begin{table}[!ht]
\centering
\caption{Prompt Template}
\vspace{2mm}
\begin{tabularx}{\linewidth}{lX}
\toprule
\textbf{} & \textbf{Template} \\ 
\midrule
Prompt Input   & Below is an instruction that describes a task, paired with an input that provides further context. Write a response that appropriately completes the request. \\
               & \\
               & \textbf{Instruction:} \{instruction\} \\
               & \\
               & \textbf{Input:} \{input\} \\
               & \\
               & \textbf{Response:} \\ 
\hdashline
\addlinespace
Prompt No Input & Below is an instruction that describes a task. Write a response that appropriately completes the request. \\
                & \\
                & \textbf{Instruction:} \{instruction\} \\
                & \\
                & \textbf{Response:} \\ 
\bottomrule
\end{tabularx}
\label{tab:prompt_template}
\vspace{-3mm}
\end{table}

\section{Additional Results}

\textbf{Additional Results for Impact of the Rescaler}. Table 5 presents the results of three variants of the scaler, where experiments were conducted in cases with 4 clients. To fully understand the impact of the scaler, we conducted additional experiments in cases with 40 clients, comparing three variants: 1) \textbf{\modelname{} with learnable rescaler} $s_i$: Our full approach with learned rescaling factors. 2) \textbf{Static rescaler} $k/k_i$: A deterministic rescaling based on the ratio between the standard number of activated experts $k$ and the client-specific reduced number $k_i$. 3) \textbf{No rescaler}: A variant without rescaling. Table \ref{tab-rescaler-app} presents these results across different datasets, data distributions, and resource budgets. Aligning with the results on 4 clients, we observe that:
\begin{packed_item}
\item \textbf{Learnable rescaler advantage}: The learnable rescaler $s_i$ generally yields the best or highly competitive performance, outperforming the "No rescaler" variant in 12 out of 16 experimental settings. For example, on Dolly with $\alpha=5$ and budget $\beta_1$, \modelname{} with $s_i$ achieves \textbf{35.25}, compared to  34.86 without a rescaler.

\item \textbf{Static rescaler limitations}: The static rescaler ($k/k_i$) consistently underperforms other variants across nearly all settings. For instance, on AlpaGasus with $\alpha=5$ and budget $\beta_4$, the static rescaler achieves only 20.35, whereas the learnable rescaler achieves 21.29, and no rescaler achieves 20.65.

\item \textbf{Resource-dependent impact}: The advantage of the learnable rescaler becomes more pronounced at certain resource levels. At $\beta_4$ (179.2B FLOPs), the learnable rescaler consistently outperforms both alternatives across all datasets and distribution settings.
\end{packed_item}

\begin{table}[!t]
\centering
\caption{Impact of the rescaler, results are reported on OLMoE-1.3B/6.9B.}
\label{tab-rescaler-app}
\scriptsize
\setlength\tabcolsep{0pt}
\begin{tabular*}{\linewidth}{@{\extracolsep{\fill}} lcccccccc}
\midrule
\multirow{3}{*}{} &\multirow{2}{*}{} &\multicolumn{3}{c}{AlpaGasus ($\alpha$ = 5)} &\multicolumn{3}{c}{AlpaGasus ($\alpha$ = 0.5)} \\
\cmidrule(l{0.00em}r{0.00em}){3-5}\cmidrule(l{0.00em}r{0.00em}){6-8}
\cmidrule(l{0.00em}r{0.00em}){3-5}\cmidrule(l{0.00em}r{0.00em}){6-8}
&FLOPs &No rescaler &static $k / k_i$ &$s_i$ (\modelname) &No rescaler &static $k / k_i$ &$s_i$ (\modelname) \\
\midrule
$\beta_4$ &153.6B &20.65 &20.35 &\textbf{21.29} &20.36 &20.14 &\textbf{20.89} \\
$\beta_3$ &179.2B &28.59 &28.28 &\textbf{29.11} &27.98 &27.70 &\textbf{29.04} \\
$\beta_2$ &230.4B &\textbf{33.87} &33.41 &33.74 &33.72 &33.22 &\textbf{34.19} \\
$\beta_1$ &332.8B &35.01 &34.53 &\textbf{35.69} &34.51 &33.86 &\textbf{35.36} \\
\midrule
\midrule
\multirow{3}{*}{} &\multirow{2}{*}{} &\multicolumn{3}{c}{Dolly ($\alpha$ = 5)} &\multicolumn{3}{c}{Dolly ($\alpha$ = 0.5)} \\
\cmidrule(l{0.00em}r{0.00em}){3-5}\cmidrule(l{0.00em}r{0.00em}){6-8}
\cmidrule(l{0.00em}r{0.00em}){3-5}\cmidrule(l{0.00em}r{0.00em}){6-8}
&FLOPs &No rescaler &static $k / k_i$ &$s_i$ (\modelname) &No rescaler &static $k / k_i$ &$s_i$ (\modelname) \\
\midrule
$\beta_4$ &153.6B &\textbf{19.81} &19.44 &19.56 &\textbf{21.32} &21.04 &21.16 \\
$\beta_3$ &179.2B &27.92 &27.40 &\textbf{29.18} &30.02 &29.52 &\textbf{30.54} \\
$\beta_2$ &230.4B &\textbf{33.74} &33.28 &33.49 &33.14 &32.66 &\textbf{34.30} \\
$\beta_1$ &332.8B &34.86 &34.48 &\textbf{35.25} &34.75 &34.34 &\textbf{34.98} \\
\midrule
\end{tabular*}
\end{table}

\begin{figure}[!t]
\centering
\begin{subfigure}[b]{0.48\textwidth}
\centering
\caption{AlpaGasus ($\alpha$ = 5)}
\includegraphics[width=0.975\textwidth]{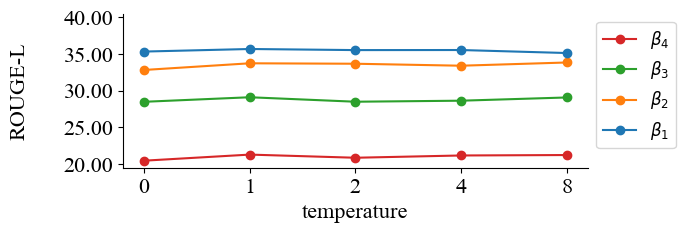}
\end{subfigure}%
\hfill
\begin{subfigure}[b]{0.48\textwidth}
\centering
\caption{AlpaGasus ($\alpha$ = 0.5)}
\includegraphics[width=0.975\textwidth]{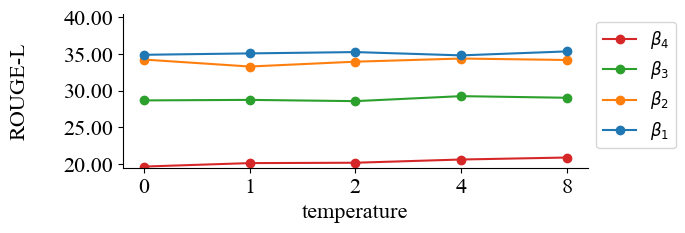}
\end{subfigure}%
\hfill
\begin{subfigure}[b]{0.48\textwidth}
\centering
\caption{Dolly ($\alpha$ = 5)}
\includegraphics[width=0.975\textwidth]{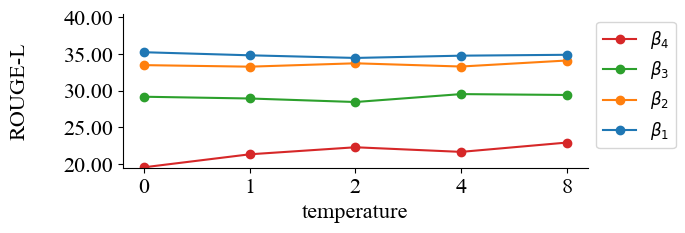}
\end{subfigure}%
\hfill
\begin{subfigure}[b]{0.48\textwidth}
\centering
\caption{Dolly ($\alpha$ = 0.5)}
\includegraphics[width=0.975\textwidth]{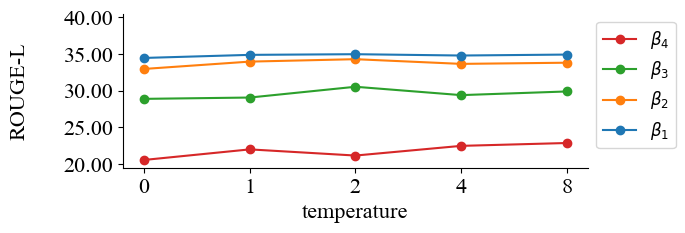}
\end{subfigure}%
\hfill
\caption{Impact of the temperature, results are reported on OLMoE-1.3B/6.9B. }
\label{fig-temperature-app}
\end{figure}

\textbf{Additional Results for Impact of Temperature in Activation-Aware Aggregation}. Figure 3 presents the results of varied temperature parameter $t$ in our activation-aware aggregation scheme, where experiments were conducted in cases with 4 clients. To fully understand the impact of the temperature, we conducted additional experiments in cases with 40 clients, comparing different temperature values ranging from t = 0 (equivalent to standard federated averaging) to t = 8 (strongly favoring high-activation clients). Figure \ref{fig-temperature-app} presents these results across different datasets, data distributions, and resource budgets. Aligning with the results on 4 clients, we observe that:

\begin{packed_item}

\item \textbf{Resource-dependent temperature sensitivity}: The most resource-constrained setting ($\beta_4$, red lines) shows the highest sensitivity to temperature, with performance consistently improving as temperature increases up to $t=4$ or $t=8$. For example, on AlpaGasus ($\alpha=0.5$), performance at $\beta_4$ increases steadily from $t=0$ to $t=8$.

\item \textbf{Dataset-specific patterns}: The Dolly dataset with $\alpha=0.5$ (high heterogeneity) shows particularly strong improvements with higher temperatures at $\beta_4$, suggesting that activation-aware aggregation is especially beneficial for resource-constrained settings with heterogeneous data distributions.

\item \textbf{Benefit of activation-aware aggregation}: Across nearly all configurations, incorporating activation frequency ($t>0$) outperforms standard federated averaging ($t=0$). This confirms our hypothesis that experts should be more heavily influenced by clients when they are frequently activated.
    
\end{packed_item}

\section{Related Work}

\textbf{Parameter-efficient Federated Fine-tuning for LLMs} represents a broad research area addressing federated learning with reduced parameter requirements. This area includes: (1) \textit{LoRA-based} methods that decompose weight updates into low-rank approximations, such as FeDeRA \cite{yan2024federa}, which uses SVD for initialization; FedSA-LoRA \cite{guo2024selective}, which selectively shares matrices; FFA-LoRA \cite{sun2024improving}, which freezes A matrices; and FEDHM \cite{yao2025fedhm}, which aggregates low-rank models into full-rank ones. FSLoRA \cite{fang2025federated} leverages a sketching mechanism to enable clients to selectively update submatrices of global LoRA modules maintained by the server. RoLoRA \cite{chen2025robust} uses alternating optimization to fine-tune LoRA adapters. (2) \textit{Prompt-based} approaches like PROMPTFL \cite{guo2023promptfl} and FedBPT \cite{sun2023fedbpt} optimize input prompts rather than model parameters. Moreover, FedPepTAO \cite{che-etal-2023-federated} chooses proper layers of prompts based on the importance of each layer, as transferring the whole set of parameters in all the prompt layers corresponds to heavy communication costs. (3) \textit{Adapter-based} methods insert specialized modules between frozen layers, including FedAdapter \cite{cai2022fedadapter}, which adjusts adapter dimensions; FedTTT \cite{ghiasvand2024communication}, which uses tensor decomposition; and C2A \cite{kim2023client}, which generates client-specific adapters via hypernetworks.

\textbf{Resource-adaptive Federated Fine-tuning}, most relevant to our work, specifically addresses heterogeneous client computational capabilities within federated learning. Existing approaches include FedIT \cite{zhang2024towards}, which applied FedAvg \cite{mcmahan2017communication} to LoRA with fixed ranks; FLoRA \cite{wang2024flora}, introducing stacking-based aggregation; HLoRA \cite{cho-etal-2024-heterogeneous}, which distributes truncated LoRA modules and uses sparsity-weighted aggregation; and FlexLoRA \cite{bai2024federated}, which leverages SVD for dynamic rank adjustment based on client resources. These methods primarily focus on compressing global LoRA matrices at different levels to accommodate diverse client capabilities, which our work identifies as fundamentally limiting for achieving true computational load adaptation. AFLoRA \cite{zhou2025aflora} decouples shared and client-specific updates to reduce overhead and improve aggregation accuracy, incorporates diag matrix-based rank pruning to better use local resources, and employs rank-aware aggregation with public data refinement to strengthen generalization under data heterogeneity. Additionally, FLoRIST \cite{ramesh2025florist} attempts to combine FLoRA and FlexLoRA to acquire better accuracy.

\textbf{Harnessing SMoE architecture}. Another related and orthogonal line of work is harnessing the SMoE architecture for diverse applications. MoEGAN \cite{chai2023improved} introduces a GAN architecture with a mixture-of-experts generator and Feature Statistics Alignment paradigm to render fine-grained learning signals to advance the generator training. QMoE \cite{frantar2023qmoe} is a new compression and execution framework, consisting of a scalable algorithm that accurately compresses trillion-parameter MoEs to less than 1 bit per parameter, in a custom format co-designed with bespoke GPU decoding kernels. LLaVA-MoLE \cite{chen2024llava} applies a sparse mixture of LoRA experts to LLaVA-1.5 \cite{liu2024improved} for instruction finetuning. Recently, several studies have preliminarily explored SMoE in federated learning settings. For example, PM-MOE \cite{feng2025pm} addresses personalized federated learning by integrating a mixture of personalized modules and an energy-based personalized module denoising, enabling each client to select beneficial personalized parameters from other clients. 
A3SMoE \cite{tran2025revisiting} preliminarily considers resource-adaptive federated learning while neglecting the MoE output divergence issue.
FLEx \cite{liu2025unlocking} aims to address excessive communication overhead by pruning the global MoE model and employs an adaptive gating mechanism to reintegrate experts into the pre-trained MoE layers.  

\section{Future Work}

Due to computational resource constraints, our experiments were limited to OLMoE-1.3B and 6.9B models. In future work, we plan to extend our evaluation to larger SMoE-based LLMs to further validate the effectiveness of our approach. Additionally, it is important to investigate the privacy robustness of our activation-aware aggregation scheme in the presence of malicious clients. Finally, we aim to develop an architecture-agnostic method for resource-adaptive federated fine-tuning of LLMs, enabling broader applicability across diverse deployment scenarios.

\end{document}